\documentclass[10pt,twocolumn]{article}

\usepackage[a4paper, margin=.8in, columnsep=.3in]{geometry}
\usepackage{times} 

\usepackage{graphicx}
\usepackage{amsmath}
\usepackage{booktabs} 

\usepackage{multirow} 
\usepackage{xspace} 
\usepackage{url}
\usepackage{hyperref}
\usepackage[compact]{titlesec}
\usepackage[numbers]{natbib}
\def\abstract{\centerline{\large\bf Abstract}\vspace*{10pt}\it}

\usepackage{fancyhdr} 
\fancypagestyle{firststyle}{
   \fancyhead[R]{\textsc{in proceedings of the 6th International Workshop on Multimedia Content Analysis in Sports (MMSports '23), October 29, 2023, Ottawa, ON, Canada.}}
}

\usepackage[]{authblk} 

\title{Context-Aware 3D Object Localization from Single Calibrated Images: A Study of Basketballs}

\author[1]{Marcello Davide Caio}
\author[1,2]{Gabriel Van Zandycke}
\author[2]{Christophe De Vleeschouwer}
\affil{
    SportRadar AG \authorcr
    {\small\nolinkurl{{m.caio.external,g.vanzandycke}@sportradar.com}}
}
\affil[2]{
    UCLouvain ICTEAM/ELEN Belgium \authorcr{\small\nolinkurl{{gabriel.vanzandycke,christophe.devleeschouwer}@uclouvain.be}}
}

\date{}
\begin{document}

\renewcommand{\floatpagefraction}{0.9}

\maketitle
\thispagestyle{firststyle} 

\begin{abstract}
Accurately localizing objects in three dimensions (3D) is crucial for various computer vision applications, such as robotics, autonomous driving, and augmented reality. This task finds another important application in sports analytics and, in this work, we present a novel method for 3D basketball localization from a single calibrated image. Our approach predicts the object's height in pixels in image space by estimating its projection onto the ground plane within the image, leveraging the image itself and the object's location as inputs. The 3D coordinates of the ball are then reconstructed by exploiting the known projection matrix. Extensive experiments on the public DeepSport dataset, which provides ground truth annotations for 3D ball location alongside camera calibration information for each image, demonstrate the effectiveness of our method, offering substantial accuracy improvements compared to recent work. Our work opens up new possibilities for enhanced ball tracking and understanding, advancing computer vision in diverse domains. The source code of this work is made publicly available at \url{https://github.com/gabriel-vanzandycke/deepsport}.
\end{abstract}

\section{Introduction}
\label{sec:intro}

Three-dimensional (3D) localization plays a crucial role in computer vision, enabling a wide range of applications across various domains. While 3D localization is relatively straightforward when certain assumptions can be made, such as objects being located on the ground plane, or when triangulation is possible, the task becomes significantly more challenging when objects are positioned off the ground and only a single image is available. Under these circumstances, this challenge extends to human perception, emphasizing the importance of developing robust computational methods for 3D localization.

\begin{figure}
  \includegraphics[width=\linewidth,trim=650 0 600 0,clip]{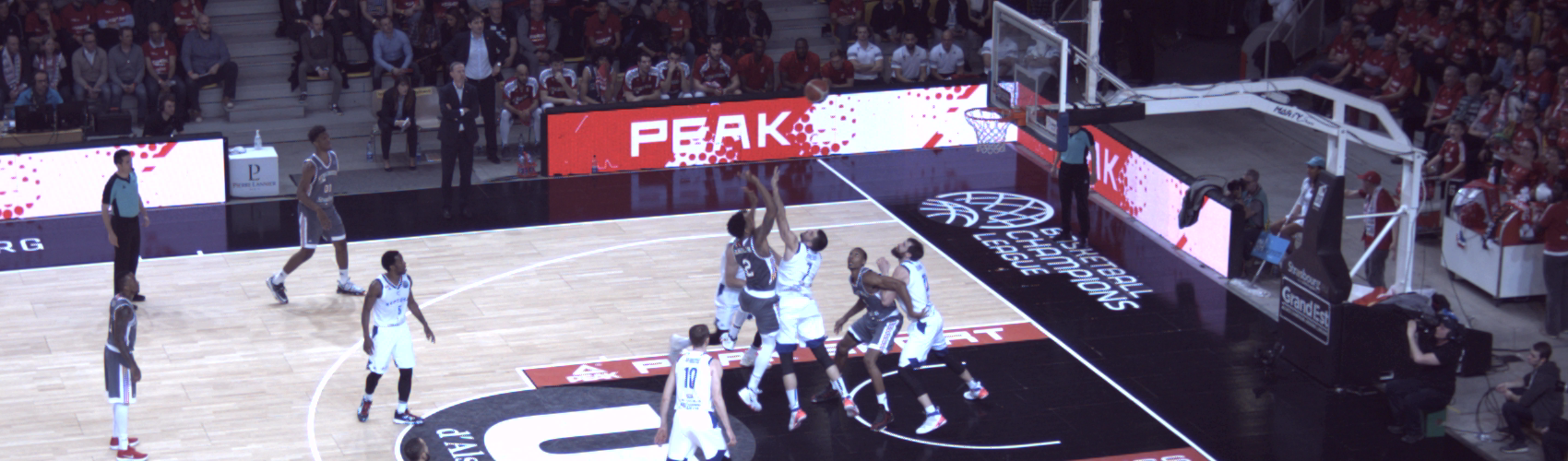}
  \caption{Can you guess the 3D location of the ball in the image? Localizing the ball in 3D from an image of intense gameplay demands insightful cues ranging from players' gaze, to the size and position of players and surrounding objects.}
  \label{fig:teaser}
\end{figure}

In sports analytics, understanding the precise 3D localization of objects, such as balls~\cite{Kamble2019} and players~\cite{Yang2018, Monezi2020}, can offer valuable insights for player performance analysis~\cite{Skinner2015, DeSilva2018}, referee assistance~\cite{Bal2012}, and audience engagement. While existing commercial solutions in sports analytics often rely on multi-camera setups~\cite{Kumar2011,Pingali2000,Ren2008,Lampert2012,Cheng2018,Parisot2011,Parisot2019,Maksai2016}, these often come with logistical and financial constraints. In scenarios where access to premium multi-camera setups is limited, the ability to leverage single calibrated camera systems for accurate ball localization becomes invaluable.
In this work, we focus on the open problem of estimating the 3D location of balls in basketball games from single calibrated images.

\begin{figure*}[t]
    \centering
    \begin{tabular}{cccc}
    \includegraphics[width=.22\linewidth]{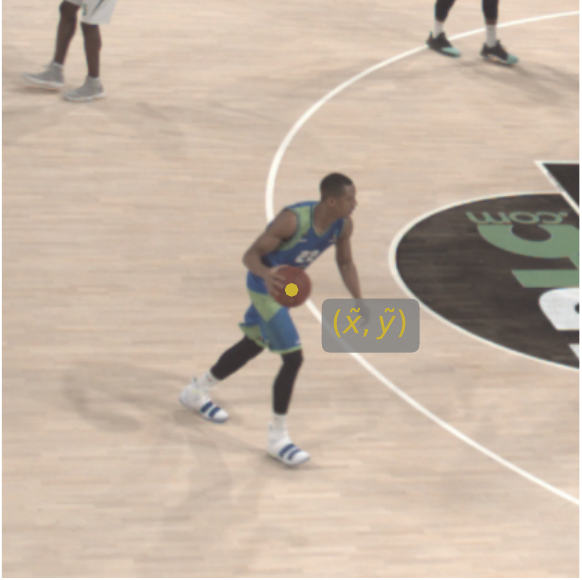} &
    \includegraphics[width=.22\linewidth]{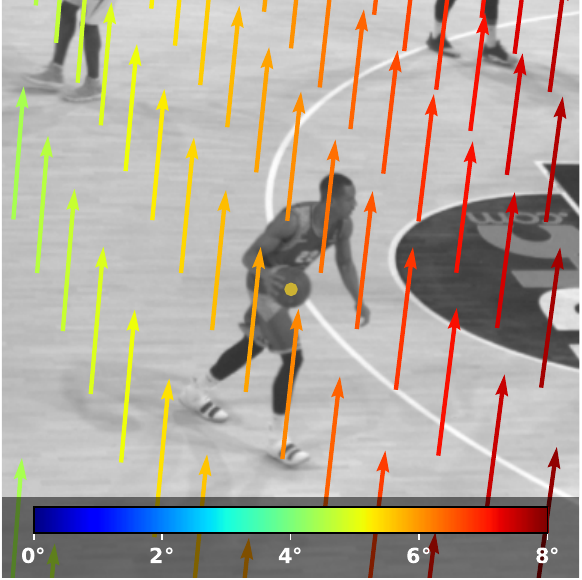} &
    \includegraphics[width=.22\linewidth]{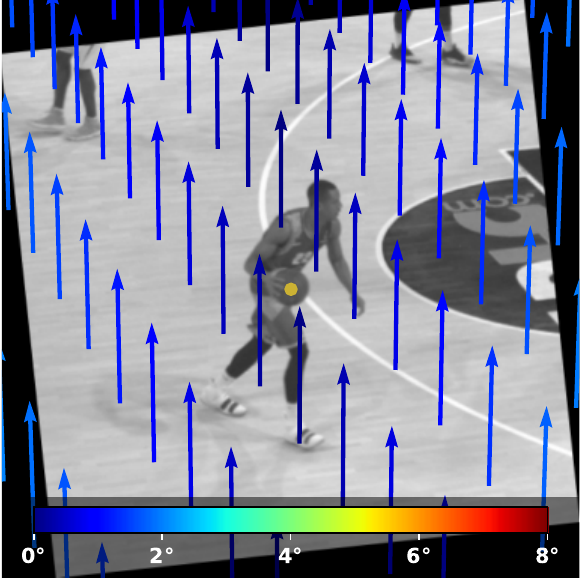} &
    \includegraphics[width=.22\linewidth]{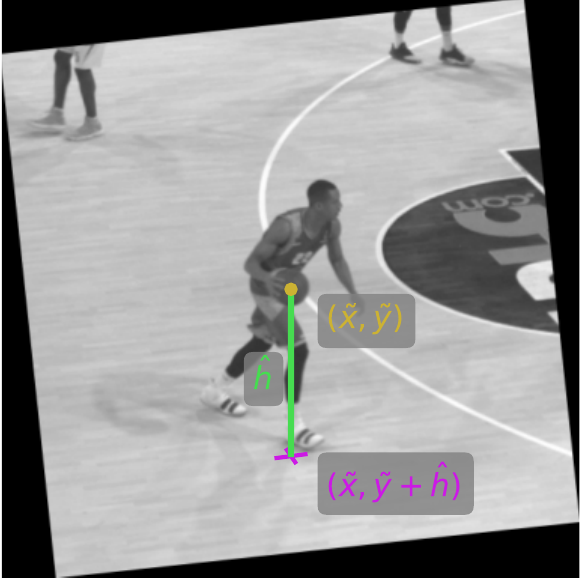} \\
    (a) & (b) & (c) & (d) \\
    \end{tabular}
    \caption{The predicted height and the given 2D position are converted to 3D coordinates using the camera calibration data. (a)~Original image with ball position $(\tilde{x}, \tilde{y})$. (b), (c)~Colored vectors perpendicular to the court floor show how verticals are oriented relative to image $y$-axis, with a scale given in degrees. (c) Image is rectified aligning the world $Z$-axis with the image $y$-axis. (d) Predicted height $\hat{h}$ is used to find the ball vertical projection on the ground floor $(\tilde{x},\tilde{y}+\hat{h})$ which can be projected to $(\tilde{X},\tilde{Y},0)$ using the calibration data. The $Z$ coordinate is obtained by re-projecting $(\tilde{x}, \tilde{y})$ to the planes $X=\tilde{X}$ or $Y=\tilde{Y}$.}
    \label{fig:method}
\end{figure*}

While 3D ball localization has been extensively studied for parabolic trajectories in calibrated videos~\cite{Chen2012,Shun2004,Chen2009,Kim1998,Skold2015,Parisot2019,Labayen2014,Silva2011}, little attention has been given to the general challenge of 3D ball localization from a single calibrated image, leaving a significant gap in research. Recent works addressing this task involved the estimation of the ball's diameter in pixels and the reconstruction of its 3D location based on this estimation\cite{ball3d,ball3dwinner}. However, this approach suffers from various limitations, including sensitivity to variations in the ball's size within the image, issues with motion blur and image resolution, and the requirement for prior knowledge of the ball's real-world size.

The observation that motivates this work is that in basketball, much like in other sports, human observers rely on contextual cues such as the position, occlusion, and size of nearby players, crowd, and court objects like basketball boards to accurately locate the ball during a game. If tasked with estimating the 3D position of the ball in a paused video, a person would use the position of the players' feet on the court as a reference for $X$ and $Y$ coordinates, while making an informed guess about the height at which the ball is held. During passes or shots, the task becomes even more challenging and cues like players gaze orientation are required to improve the height perception. The idea is to give enough context to the model to guess the ball height in image space, and use geometry to retrieve the 3D ball coordinates.

This paper introduces a novel approach to 3D ball localization in basketball from single calibrated images. Building upon previous work~\cite{VanZandycke2019, Ghasemzadeh2021}, we assume that the ball has already been detected in the image, allowing us to crop the image region around the ball. By providing sufficient context within the crop, our model predicts the height of the ball in image space, measured in pixels. Furthermore, by placing the ball at the center of the crop, we implicitly convey its location to the model. 

Besides lifting the limitations of the previous method, this approach is more versatile and easily transferable to other sports such as soccer, volleyball, handball, and potentially extensible to domains beyond sports.

\section{Method}
\label{sec:method}

We propose to estimate the 3D location of a ball by training a neural network model to predict the distance in pixels between the ball and its ground projection in image space. Retrieving the 3D ball coordinates is then easily achieved by leveraging the projection matrix from the camera calibration data.

\subsection{Ball height estimation in pixels}
\label{sec:model}
Our model has a Convolutional Neural Network (CNN) backbone to extract image features, which are subsequently processed by a regression head supervised to predict the distance in pixels between the ball and its ground projection in the image space.
The model operates on square image crops centered on the ball, ensuring sufficient contextual information is provided.

\subsection{Pixels height to world coordinates}
\label{sec:2dto3d}

\begin{figure*}
    \centering
\begin{tabular}{c@{\hskip 4.5pt}c@{\hskip 4.5pt}c@{\hskip 4.5pt}c@{\hskip 4.5pt}c}
\includegraphics[width=0.032\linewidth]{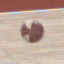}&
\includegraphics[width=0.064\linewidth]{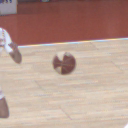}&
\includegraphics[width=0.128\linewidth]{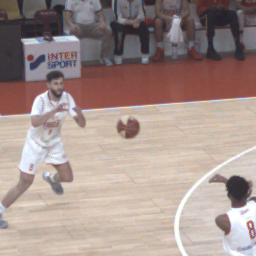}&
\includegraphics[width=0.256\linewidth]{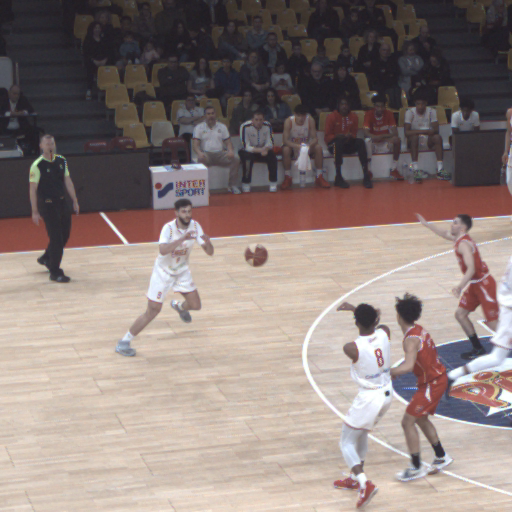}&
\includegraphics[width=0.4\linewidth]{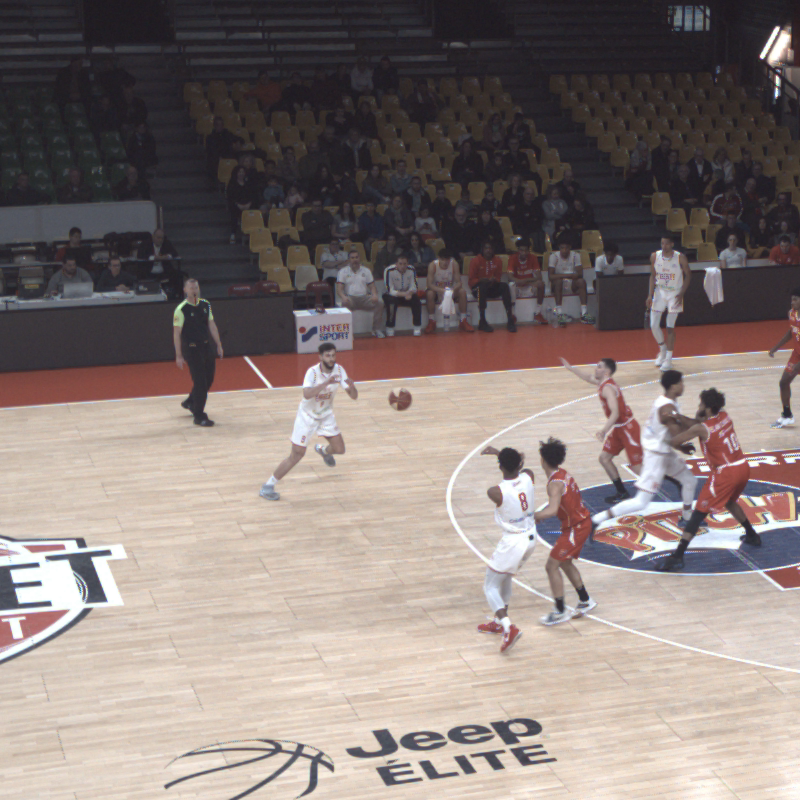}
\\
$64\times64$&$128\times128$&$256\times256$&$512\times512$&$800\times800$
\end{tabular}
    \caption{Various crop sizes for a given image in the dataset. Larger crop sizes offer more contextual information to the model.}
    \label{fig:crops}
\end{figure*}

Let us define a 3D coordinates system $\left(X,Y,Z\right)$ with the horizontal ground plane defined by $Z=0$, for the real world, and a 2D coordinate system $(x,y)$, for image space. Details of the camera calibration allow us to project 3D points from the real world to the image space, and to convert 2D points in image space into light rays in 3D. To transform the estimated height of the ball in image space into its corresponding 3D coordinates in the real world, we apply the following procedure; see Fig.~\ref{fig:method}.

First, the input image is undistorted, ensuring that straight lines in the real world correspond to straight lines in image space. Then, the image is rotated, aligning the $Z$-axis in the real world with the vertical direction $y$ in the image space. 
After rotation, the predicted height of the ball in pixels $\hat{h}$ effectively represents the vertical displacement between the ball's position in the image $(\tilde x, \tilde y)$, located at the center of the crop, and its projection onto the ground in image space $(\tilde x, \tilde y + \hat{h})$.
As the camera is calibrated, we can project $(\tilde x, \tilde y + \hat h)$ into 3D space on the $Z=0$ ground plane, determining the position of the ball's projection on the ground $(\tilde X, \tilde Y, 0)$. Finally, to obtain the 3D world coordinates of the ball, we perform a reprojection process. We trace the path of the light ray passing through the ball in the image space, determining the two 3D points at which it intersects the $Y=\tilde Y$ and $X=\tilde X$ vertical planes. By taking the average of these two points, we retrieve the 3D ball coordinates.

\section{Datasets}
\label{sec:dataset}

In this study, we aim to evaluate the effectiveness of our proposed method for ball 3D localization in basketball. To accomplish this, we require a suitable dataset that provides high-resolution images with accurate ball annotations. While datasets such as Basket-APIDIS~\cite{Vleeschouwer2008} and Soccer-ISSIA~\cite{DOrazio2009} have been previously utilized for related tasks, we exclude them from consideration as the concerns raised in reference \cite{ball3d} also apply to this work.

\paragraph{DeepSport dataset}
Following Ref.~\cite{deepsportradar}, we utilize for training of our models the public DeepSport dataset~\cite{DeepSportDataset, deepsport-dataset}, introduced in Ref.~\cite{VanZandycke2019} and expanded in Ref.~\cite{deepsportradar}, which offers 364 high-resolution panoramic images captured during professional basketball matches across 15 different arenas. The images in the dataset vary in resolution, typically reaching approximately $4500\textstyle{\times}1500$ pixels. The ball's 3D annotation is provided through the annotation of its center and vertical projection on the court within the image space.

\paragraph{Ballistic dataset}
Beside evaluation of our model on the DeepSport dataset, in Section \ref{sec:benchmark-ballistic}, we benchmark it also on the evaluation dataset introduced in Ref.~\cite{ball3d} that features highly reliable 3D ball annotations from ballistic trajectories. This dataset is composed of $233$ images of $2336\textstyle\times1756$ pixels resolution, coming from $2$ different basketball arenas.

\section{Experiments}
\label{sec:experiments}


\subsection{Experiment configurations}
Implementation details and metrics used to conduct this research are presented hereafter.

\paragraph{Dataset}
As mentioned in Section~\ref{sec:dataset}, we use the DeepSport dataset for training, and have adopted the split defined in~\cite{DeepSportDataset}, where each fold contains images from basketball arenas exclusive to that fold. The fold \texttt{A} remained unseen and was used for testing. 

\paragraph{Baseline}
As a baseline for our approach, we use a model introduced in Ref.~\cite{ball3d} whereby the ball 3D position is obtained by estimating its diameter in the image space and combining knowledge of real ball size with calibration data.

\paragraph{Architecture} 
We study the impact of the feature extractor by experimenting with several popular architectures, including VGG, ResNet, MobileNet and EfficientNet; all pre-trained on ImageNet~\cite{Deng2010} from the TensorFlow library~\cite{tensorflow}.
The regression head, borrowed from Ref.~\cite{ball3d}, contains 3 layers, and outputs a scalar supervised to predict the ball height in pixels.

\paragraph{Context size}
We thoroughly investigate the impact of the square crop size on our model's performance. It is evident that a small crop around the basketball would lack sufficient context for the model to accurately estimate its projection onto the ground. By enlarging the crop size and incorporating elements from the surrounding scene, we anticipate a significant enhancement in model performance. However, it is important to note that as the crop size increases, the improvements in performance may eventually plateau, leading to diminishing returns.

\paragraph{Image scale}
In our research, we investigate the influence of the input image scale on our model's performance, as it directly affects the representation of the basketball and all relevant objects and people within the scene.
Moreover, the combined effect of image scale and context size also influences the inference speed of the model, a critical consideration for real-time analysis in production systems.

\paragraph{Training procedure}
The backbone is pre-trained on ImageNet~\cite{Deng2010} from the TensorFlow library~\cite{tensorflow}, and we initialize the regression head with the Glorot initializer~\cite{Glorot2010}.
The neural network is then trained on image patches centered around the oracle ball positions for 100 epochs using Adam optimizer~\cite{Kingma2015} with a learning rate of $10^{-4}$. The final height prediction is supervised with a Huber loss~\cite{Huber1964} with $\delta=1.0$. Each experiment is repeated with 8 different initialization of the regression head and every reported metric is composed of mean and standard deviation computed over the 8 repetitions.

\paragraph{Metrics}
We evaluate the performance of our 3D ball localization method using the following metrics:
\begin{itemize}
    \item Mean Absolute Error (MAE [px]): This metric measures the average error in pixels for the predicted height of the ball in image space.
    \item Mean and Median Absolute Projection Error (MAPE [m] and MdnAPE [m]): These metrics quantify the average and median error in meters for the predicted projection of the ball on the court floor, with respect to the ground truth basketball projection.
    \item Mean and Median Absolute 3D Error (MA3DE [m] and MdnA3DE [m]): These metrics assess the average and median error in meters for the reconstructed 3D ball position, relative to the ground truth 3D ball position.
\end{itemize}
By utilizing these metrics, we can comprehensively assess the accuracy and performance of our proposed 3D ball localization method.
As the metrics are computed on oracle ball positions in image space, they reveal the true errors from our proposed approach, decorrelated from any error that a detector upstream could introduce.

\subsection{Evaluation against the baseline}
As a first experiment, we validate our proposed method for 3D ball localization against the model presented in Ref.~\cite{ball3d}, which relied on ball size estimation. For our model, we select a reasonable set of hyperparameters: the crop size is fixed to $512$ pixels (about $1/3$rd of the average dataset image height), the backbone is VGG16, and the image resolution is the one of the dataset. These choice of hyperparameters will be taken as default in the following sections. To ensure a fair comparison, we retrained the baseline model using the same hyperparameters as described in the previous paper and ran the experiments under similar conditions.

\begin{table}[ht]
\centering
\begin{tabular}{lcc}
\toprule
\textbf{Metrics} & \textbf{Our Method} & \textbf{Baseline}\\
\midrule
MAE [px] & $34\pm3$ & N/A\\
MAPE [m] & $1.25\pm0.11$ & $2.88\pm0.22$\\
MdnAPE [m] & $0.92\pm0.15$ & $2.10\pm0.38$\\
MA3DE [m] & $1.29\pm0.10$ & $2.97\pm0.23$\\
MdnA3DE [m] & $0.95\pm0.16$ & $2.18\pm0.40$\\
\bottomrule
\end{tabular}
\caption{Comparison of performance metrics between the proposed method and baseline model~\cite{ball3d}.}
\label{tab:experiment1_results}
\end{table}

The results in Table~\ref{tab:experiment1_results} demonstrate the superiority of our proposed method over the baseline model. The significant improvement in MAPE and MA3DE validates the effectiveness of our approach for accurate 3D ball localization, setting a solid foundation for further experimentation and analysis.

The notable discrepancy between median and mean values observed across all experiments, including both our proposed method and the baseline, indicates that current solutions for ball 3D localization are significantly affected by outliers. Consequently, the performance of these models in production may be better than what is suggested by mean value of metrics. 

\subsection{Impact of backbone}
We considered several popular backbone architectures including VGG, ResNet, MobileNet and EfficientNet. Our findings show that they all have similar performance when trained on the DeepSport dataset. We anticipate that when exposed to a larger dataset, larger models would perform better, at the cost of longer inference time.

\subsection{Impact of context size}

Here, we investigate the influence of context size on the predictive power of our model for 3D ball localization; see Fig.~\ref{fig:crops}. Smaller crop sizes provide faster inference and require fewer resources, but may lack sufficient context for accurate height estimation. Conversely, larger crop sizes offer more contextual information, potentially improving model performance. Our goal is to identify the ideal crop size for a production environment where both performance and speed are important. We explore a range of crop sizes: $64$, $96$, $128$, $256$, $320$, $480$, $512$, $640$, and $800$ pixels. For each crop size, we report in Table~\ref{tab:experiment2_results} the resulting MAPE.

\begin{table}[ht]
\centering
\begin{tabular}{ccc}
\toprule
\textbf{Crop Size (px)} & \textbf{MAPE [m]} & \textbf{MA3DE [m]} \\
\midrule
$64$ & $1.71\pm0.17$   & $1.75\pm0.17$ \\
$96$ & $1.52\pm0.13$   & $1.56\pm0.13$ \\
$128$ & $1.46\pm0.06$  & $1.50\pm0.06$ \\
$256$ & $1.04\pm0.11$  & $1.07\pm0.12$ \\
$320$ & $1.17\pm0.09$  & $1.20\pm0.10$ \\
$480$ & $1.18\pm0.07$  & $1.21\pm0.07$ \\
$512$ & $1.25\pm0.11$  & $1.22\pm0.11$ \\
$640$ & $1.16\pm0.07$  & $1.20\pm0.08$ \\
$800$ & $1.08\pm0.06$  & $1.11\pm0.06$ \\
\bottomrule
\end{tabular}
\caption{Impact of context size on MAPE. The crop size is varied whilst keeping the image scale fixed.}
\label{tab:experiment2_results}
\end{table}

As expected, a small crop size such as $64\times64$ yields a higher MAPE, indicating limited context for accurate height estimation. As the crop size increases, MAPE decreases, indicating improved performance. However, MAPE already plateaus beyond a crop size of $256$ pixels. We thus identify the crop size of 256 pixels as the optimal choice, striking a balance between model performance and efficient inference, with promising implications for real-time applications.

\subsection{Impact of image scale}\label{sec:image-res}
In this section, we present a comprehensive study to assess the robustness of our model with respect to input image scale. The dataset comprises panoramic images of the basketball court. To evaluate the impact of image scale, effectively we shrink the images at various ratios ($1$, $1/2$, $1/4$, and $1/8$). The square crop used as input to the model is also scaled accordingly, such that the model has access to the same contextual information independently of the scale.

\begin{table}[ht]
\centering
\begin{tabular}{cc}
\toprule
\textbf{Scale Ratio} & \textbf{MAPE [m]} \\
\midrule
$1$ & $1.25\pm0.11$ \\
$1/2$ & $1.33\pm0.11$ \\
$1/4$ & $1.32\pm0.08$ \\
$1/8$ & $1.22\pm0.07$ \\
\bottomrule
\end{tabular}
\caption{Performance of the model with respect to input image size.}
\label{tab:experiment3_results}
\end{table}

From the results in Table \ref{tab:experiment3_results}, we observe that our proposed method exhibits remarkable robustness concerning image resolution. Despite shrinking the images up to $1/8$th of the original size, the model's performance remains consistent. This robustness stands in stark contrast to the approach presented in Ref.~\cite{ball3d}, where high-resolution images were a prerequisite for achieving accurate ball localization.

\subsection{Benchmark on ballistic dataset}\label{sec:benchmark-ballistic}
\begin{figure}[ht]
  \centering
  \includegraphics[width=.7\linewidth]{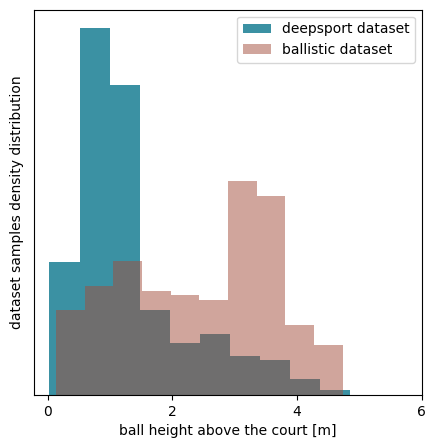}
  \caption{Ball height distribution in meters for the DeepSport dataset (training and evaluation sets combined), and the ballistic dataset introduced in Ref.~\cite{ball3d}.}
  \label{fig:height-distribution}
\end{figure}

We evaluate our method on a benchmark dataset of balls moving on ballistic trajectories, as introduced in Ref.\cite{ball3d}. The obtained Mean Absolute Projection Error (MAPE) on this dataset is $2.68\pm0.12$ m, significantly higher than the results obtained on the DeepSport test set; see Table\ref{tab:experiment1_results}. The poor performance on the ballistic dataset is promptly explained by observing the distinct distribution of ball heights in the two datasets; see Fig.~\ref{fig:height-distribution}. The ballistic dataset shows a skew towards higher ball heights, with nearly half of the samples (102/233) having balls above $3$ m, while the DeepSport dataset primarily comprises balls below this mark, with only 60 of 801 training samples reaching $3$ meters.

To mitigate this limitation, we attempted to balance the number of balls above and below $2$ meters in the training set. This approach resulted in an improved performance on the ballistic dataset, reducing the MAPE to $2.21\pm0.27$ m. Nonetheless, the limited number of high-ball samples in the training dataset remains a problem: feeding the same few high-ball samples to the neural network many times during training can not help much the model to generalise to those situations. We anticipate that a larger dataset, featuring more samples with high-ball instances, could narrow the gap between the performance on the ballistic dataset and the DeepSport test set.

\section{Discussion}
In this paper, we introduced a novel method for 3D ball localization, achieving impressive results with MAPE as low as 1.1 meters. Notably, our proposed approach exhibits remarkable robustness with respect to image resolution, allowing the use of low-resolution images with the same crop size, thereby providing more context while maintaining good performance and inference speed.

Although our method demonstrates significant improvements in various scenarios, it underperforms on the ballistic dataset~\cite{ball3d}. However, this outcome is not surprising given its skewed distribution of ball heights, which significantly differs from the one on which our models were trained.

While our method is based on single images, there is potential for even better performance when applied to videos, by means of temporal smoothing or tracking. However, this goes beyond the scopes of this paper and would require an ad-hoc dataset, at least for evaluation.

Lastly, the performance of our approach could be further optimized with a more extensive and diverse training dataset. With additional data, our approach has the potential to achieve even better results, and the use of larger backbones may yield higher accuracy.

\section{Conclusion}
In conclusion, we have presented a novel method for 3D ball localization that offers a significant improvement over existing approaches in terms of accuracy and robustness. Importantly, this approach can be readily extended to other sports, given the creation of similar datasets tailored to each sport.

Moreover, our method's underlying concept of height detection in the image space and subsequent 3D position reconstruction holds promise for general 3D object localization. While each application may adapt the idea to suit specific tasks, the fundamental framework remains applicable.
\label{sec:concl}

\section*{Acknowledgements}
Part of this work has been funded by the Walloon Region project DeepSport and by Sportradar AG.
C.~De~Vleeschouwer is a Research Director of the Fonds de la Recherche Scientifique - FNRS.
Computational resources have been provided by the supercomputing facilities of the Université catholique de Louvain (CISM/UCL) and the Consortium des Équipements de Calcul Intensif en Fédération Wallonie Bruxelles (CÉCI) funded by the Fond de la Recherche Scientifique de Belgique (F.R.S.-FNRS) under convention 2.5020.11 and by the Walloon Region. M.~D.~Caio and G. Van Zandycke thank Sportradar AG for enabling and sponsoring this work, and for its commitment to scientific research, which made it possible for us to make significant contributions to the sports technology field. We also thank Davide Zambrano and the rest of the Computer Vision and Deep Learning team at Sportradar AG for their overall support of this work and for the fruitful conversations that led to the findings presented in this paper.

\bibliographystyle{plain}
\bibliography{ball3d}

\begin{thebibliography}{10}

\bibitem{Bal2012}
Baljinder Bal and Gaurav Dureja.
\newblock {Hawk Eye: A Logical Innovative Technology Use in Sports for
  Effective Decision Making}.
\newblock {\em Sport Science Review}, 21(1-2):107--119, may 2012.

\bibitem{Chen2009}
Hua~Tsung Chen, Ming~Chun Tien, Yi~Wen Chen, Wen~Jiin Tsai, and Suh~Yin Lee.
\newblock {Physics-based ball tracking and 3D trajectory reconstruction with
  applications to shooting location estimation in basketball video}.
\newblock {\em Journal of Visual Communication and Image Representation},
  20(3):204--216, 2009.

\bibitem{Chen2012}
Hua~Tsung Chen, Wen~Jiin Tsai, Suh~Yin Lee, and Jen~Yu Yu.
\newblock {Ball tracking and 3D trajectory approximation with applications to
  tactics analysis from single-camera volleyball sequences}.
\newblock {\em Multimedia Tools and Applications}, 60(3):641--667, 2012.

\bibitem{Cheng2018}
Xina Cheng, Norikazu Ikoma, Masaaki Honda, and Takeshi Ikenaga.
\newblock {Simultaneous physical and conceptual ball state estimation in
  volleyball game analysis}.
\newblock {\em 2017 IEEE Visual Communications and Image Processing, VCIP
  2017}, 2018-Janua:1--4, 2018.

\bibitem{DeSilva2018}
Varuna De~Silva, Mike Caine, James Skinner, Safak Dogan, Ahmet Kondoz, Tilson
  Peter, Elliott Axtell, Matt Birnie, and Ben Smith.
\newblock Player tracking data analytics as a tool for physical performance
  management in football: A case study from chelsea football club academy.
\newblock {\em Sports}, 6(4), 2018.

\bibitem{Deng2010}
Jia Deng, Wei Dong, Richard Socher, Li-Jia Li, {Kai Li}, and {Li Fei-Fei}.
\newblock {ImageNet: A large-scale hierarchical image database}.
\newblock {\em Institute of Electrical and Electronics Engineers (IEEE)}, pages
  248--255, mar 2010.

\bibitem{DOrazio2009}
T.~D'Orazio, M.~Leo, N.~Mosca, P.~Spagnolo, and P.~L. Mazzeo.
\newblock {A semi-automatic system for ground truth generation of soccer video
  sequences}.
\newblock In {\em 6th IEEE International Conference on Advanced Video and
  Signal Based Surveillance, AVSS 2009}, pages 559--564, 2009.

\bibitem{Ghasemzadeh2021}
Seyed~Abolfazl Ghasemzadeh, Gabriel~Van Zandycke, Maxime Istasse, Niels Sayez,
  Amirafshar Moshtaghpour, and Christophe~De Vleeschouwer.
\newblock Deepsportlab: a unified framework for ball detection, player instance
  segmentation and pose estimation in team sports scenes.
\newblock In {\em 32nd British Machine Vision Conference 2021, {BMVC} 2021,
  Online, November 22-25, 2021}, page 379. {BMVA} Press, 2021.

\bibitem{Glorot2010}
Xavier Glorot and Yoshua Bengio.
\newblock {Understanding the difficulty of training deep feedforward neural
  networks}.
\newblock In {\em Journal of Machine Learning Research}, volume~9, pages
  249--256, 2010.

\bibitem{Huber1964}
Peter~J. Huber.
\newblock {Robust Estimation of a Location Parameter}.
\newblock {\em The Annals of Mathematical Statistics}, mar 1964.

\bibitem{Kamble2019}
Paresh~R. Kamble, Avinash~G. Keskar, and Kishor~M. Bhurchandi.
\newblock {Ball tracking in sports: a survey}.
\newblock {\em Artificial Intelligence Review}, 52(3):1655--1705, 2019.

\bibitem{Kim1998}
Taeone Kim, Yongduek Seo, and Ki~Sang Hong.
\newblock {Physics-based 3D position analysis of a soccer ball from monocular
  image sequences}.
\newblock {\em Proceedings of the IEEE International Conference on Computer
  Vision}, pages 721--726, 1998.

\bibitem{Kingma2015}
Diederik~P. Kingma and Jimmy Ba.
\newblock Adam: {A} method for stochastic optimization.
\newblock In Yoshua Bengio and Yann LeCun, editors, {\em 3rd International
  Conference on Learning Representations, {ICLR} 2015, San Diego, CA, USA, May
  7-9, 2015, Conference Track Proceedings}, 2015.

\bibitem{Kumar2011}
K.~C.~Amit Kumar, Pascaline Parisot, and Christophe {De Vleeschouwer}.
\newblock {Demo: Spatio-temporal template matching for ball detection}.
\newblock In {\em 2011 5th ACM/IEEE International Conference on Distributed
  Smart Cameras, ICDSC 2011}, 2011.

\bibitem{Labayen2014}
Mikel Labayen, Igor~G. Olaizola, Naiara Aginako, and Julian Florez.
\newblock {Accurate ball trajectory tracking and 3D visualization for
  computer-assisted sports broadcast}.
\newblock {\em Multimedia Tools and Applications}, 73(3):1819--1842, dec 2014.

\bibitem{Lampert2012}
Christoph~H. Lampert and Jan Peters.
\newblock {Real-time detection of colored objects in multiple camera streams
  with off-the-shelf hardware components}.
\newblock {\em Journal of Real-Time Image Processing}, 7, 2012.

\bibitem{ball3dwinner}
Adrien Maglo, Astrid Orcesi, and Quoc-Cuong Pham.
\newblock A resnet-18 network to estimate the ball diameter in basketball
  images, 2022.

\bibitem{Maksai2016}
Andrii Maksai, Xinchao Wang, and Pascal Fua.
\newblock {What players do with the ball: A physically constrained interaction
  modeling}.
\newblock In {\em Proceedings of the IEEE Computer Society Conference on
  Computer Vision and Pattern Recognition}, volume 2016-Decem, pages 972--981,
  2016.

\bibitem{Monezi2020}
Lucas~Antônio Monezi, Anderson Calderani~Junior, Luciano~Allegretti
  Mercadante, Leonardo~Tomazeli Duarte, and Milton~S. Misuta.
\newblock A video-based framework for automatic 3d localization of multiple
  basketball players: A combinatorial optimization approach.
\newblock {\em Frontiers in Bioengineering and Biotechnology}, 8, 2020.

\bibitem{Parisot2011}
Pascaline Parisot and Christophe {De Vleeschouwer}.
\newblock {Graph-based filtering of ballistic trajectory}.
\newblock In {\em Proceedings - IEEE International Conference on Multimedia and
  Expo}, 2011.

\bibitem{Parisot2019}
Pascaline Parisot and Christophe {De Vleeschouwer}.
\newblock {Consensus-based trajectory estimation for ball detection in
  calibrated cameras systems}.
\newblock {\em Journal of Real-Time Image Processing}, 2019.

\bibitem{Pingali2000}
Gopal Pingali, Agata Opalach, and Yves Jean.
\newblock {Ball tracking and virtual replays for innovative tennis broadcasts}.
\newblock {\em Proceedings-International Conf. on Pattern Recognition}, 15(4),
  2000.

\bibitem{Ren2008}
Jinchang Ren, James Orwell, Graeme~A. Jones, and Ming Xu.
\newblock {Real-time modeling of 3-D soccer ball trajectories from multiple
  fixed cameras}.
\newblock {\em IEEE Transactions on Circuits and Systems for Video Technology},
  18(3):350--362, 2008.

\bibitem{Shun2004}
Hubert Shum and Taku Komura.
\newblock {A spatiotemporal approach to extract the 3D trajectory of the
  baseball from a single view video sequence}.
\newblock {\em 2004 IEEE International Conference on Multimedia and Expo
  (ICME)}, 3:1583--1586, 2004.

\bibitem{Silva2011}
Hugo Silva, Andr{\'{e}} Dias, Jos{\'{e}} Almeida, Alfredo Martins, and Eduardo
  Silva.
\newblock {Real-Time 3D Ball Trajectory Estimation for RoboCup Middle Size
  League Using a Single Camera}.
\newblock {\em Lecture Notes in Computer Science (including subseries Lecture
  Notes in Artificial Intelligence and Lecture Notes in Bioinformatics)}, 7416
  LNCS:586--597, 2011.

\bibitem{Skinner2015}
Brian Skinner and Stephen~J. Guy.
\newblock A method for using player tracking data in basketball to learn player
  skills and predict team performance.
\newblock {\em PLOS ONE}, 10(9):1--15, 09 2015.

\bibitem{Skold2015}
Jonas Sk{\"{o}}ld.
\newblock {Estimating 3D-trajectories from Monocular Video Sequences Estimating
  3D-trajectories from Monocular Video Sequences}.
\newblock {\em KTH Royal Institute of Technology}, 2015.

\bibitem{tensorflow}
TensorFlow.
\newblock Tensorflow library, feb 2022.

\bibitem{deepsport-dataset}
Gabriel {Van Zandycke}.
\newblock {Image and Signal Processing Group (UCL) | Softwares:
  \url{https://sites.uclouvain.be/ispgroup/Softwares/DeepSport}}, May 2020.

\bibitem{DeepSportDataset}
Gabriel {Van Zandycke}.
\newblock {DeepSport dataset:
  \url{https://www.kaggle.com/gabrielvanzandycke/deepsport-dataset}}, 2021.

\bibitem{VanZandycke2019}
Gabriel {Van Zandycke} and Christophe {De Vleeschouwer}.
\newblock {Real-time CNN-based segmentation architecture for ball detection in
  a single view setup}.
\newblock {\em MMSports 2019 - Proceedings of the 2nd International Workshop on
  Multimedia Content Analysis in Sports, co-located with MM 2019}, pages
  51--58, 2019.

\bibitem{ball3d}
Gabriel {Van Zandycke} and Christophe {De Vleeschouwer}.
\newblock {3D Ball Localization From A Single Calibrated Image}, 2022.

\bibitem{deepsportradar}
Gabriel Van~Zandycke, Vladimir Somers, Maxime Istasse, Carlo~Del Don, and
  Davide Zambrano.
\newblock Deepsportradar-v1: Computer vision dataset for sports understanding
  with high quality annotations.
\newblock In {\em Proceedings of the 5th International ACM Workshop on
  Multimedia Content Analysis in Sports}, MMSports '22, page 1–8, New York,
  NY, USA, 2022. Association for Computing Machinery.

\bibitem{Vleeschouwer2008}
Christophe~De Vleeschouwer, Fan Chen, and Damien Delannay.
\newblock {Distributed video acquisition and annotation for sport-event
  summarization}.
\newblock {\em NEM summit}, 2008.

\bibitem{Yang2018}
Yukun Yang, Min Xu, Wanneng Wu, Ruiheng Zhang, and Yu~Peng.
\newblock 3d multiview basketball players detection and localization based on
  probabilistic occupancy.
\newblock In {\em 2018 Digital Image Computing: Techniques and Applications
  (DICTA)}, pages 1--8, 2018.

\end{thebibliography}

\end{document}